\documentclass[11pt,letterpaper]{article}

\usepackage[margin=1in]{geometry}
\usepackage[T1]{fontenc}
\usepackage[utf8]{inputenc}
\usepackage{lmodern}
\usepackage{microtype}

\usepackage{nameref}
\usepackage{graphicx}
\usepackage{subcaption}
\usepackage{booktabs}
\usepackage{tabularx}
\usepackage{multicol}
\usepackage{caption}
\usepackage{multirow}
\usepackage{makecell}
\usepackage{amsmath}
\usepackage{float}
\usepackage{array}
\usepackage{pifont}
\usepackage[hyphens]{url}
\usepackage{hyperref}
\usepackage[numbers]{natbib}
\usepackage{algorithm}
\usepackage{algorithmic}
\usepackage{newfloat}
\usepackage{listings}
\usepackage{bibentry}

\newcommand{\cmark}{\ding{51}}
\newcommand{\xmark}{\ding{55}}

\DeclareCaptionStyle{ruled}{labelfont=normalfont,labelsep=colon,strut=off}
\floatstyle{ruled}
\newfloat{listing}{tb}{lst}{}
\floatname{listing}{Listing}

\title{Pharos-ESG: A Framework for Multimodal Parsing, Contextual Narration, and Hierarchical Labeling of ESG Report}

\author{
\large
Yan Chen$^{1,*}$\thanks{Corresponding author: chen\_yan@swufe.edu.cn} \qquad
Yu Zou$^{1,*}$ \qquad
Jialei Zeng$^{1}$ \\[0.3em]
Haoran You$^{1}$ \qquad
Xiaorui Zhou$^{1}$ \qquad
Aixi Zhong$^{1}$ \\[0.9em]
\small
$^{1}$Southwestern University of Finance and Economics, Chengdu, China \\[0.2em]
$^{*}$Equal contribution
}

\date{}

\begin{document}

\maketitle
\begin{abstract}

Environmental, Social, and Governance (ESG) principles are reshaping the foundations of global financial governance, transforming capital allocation architectures, regulatory frameworks, and systemic risk coordination mechanisms. 
However, as the core medium for assessing corporate ESG performance, the ESG reports present significant challenges for large-scale understanding, due to chaotic reading order from slide-like irregular layouts and implicit hierarchies arising from lengthy, weakly structured content.
To address these challenges, we propose \textbf{Pharos-ESG}, a unified framework that transforms ESG reports into structured representations through multimodal parsing, contextual narration, and hierarchical labeling.
It integrates a reading-order modeling module based on layout flow, hierarchy-aware segmentation guided by table-of-contents anchors, and a multimodal aggregation pipeline that contextually transforms visual elements into coherent natural language.
The framework further enriches its outputs with ESG, GRI, and sentiment labels, yielding annotations aligned with the analytical demands of financial research.
Extensive experiments on annotated benchmarks demonstrate that Pharos-ESG consistently outperforms both dedicated document parsing systems and general-purpose multimodal models. 
In addition, we release \textbf{Aurora-ESG}, the first large-scale public dataset of ESG reports, spanning Mainland China, Hong Kong, and U.S. markets, featuring unified structured representations of multimodal content, enriched with fine-grained layout and semantic annotations to better support ESG integration in financial governance and decision-making.
\end{abstract}

\section{Introduction}




\begin{figure}[ht]
  \centering

  \begin{subfigure}[b]{0.52\columnwidth}
    \includegraphics[width=\linewidth]{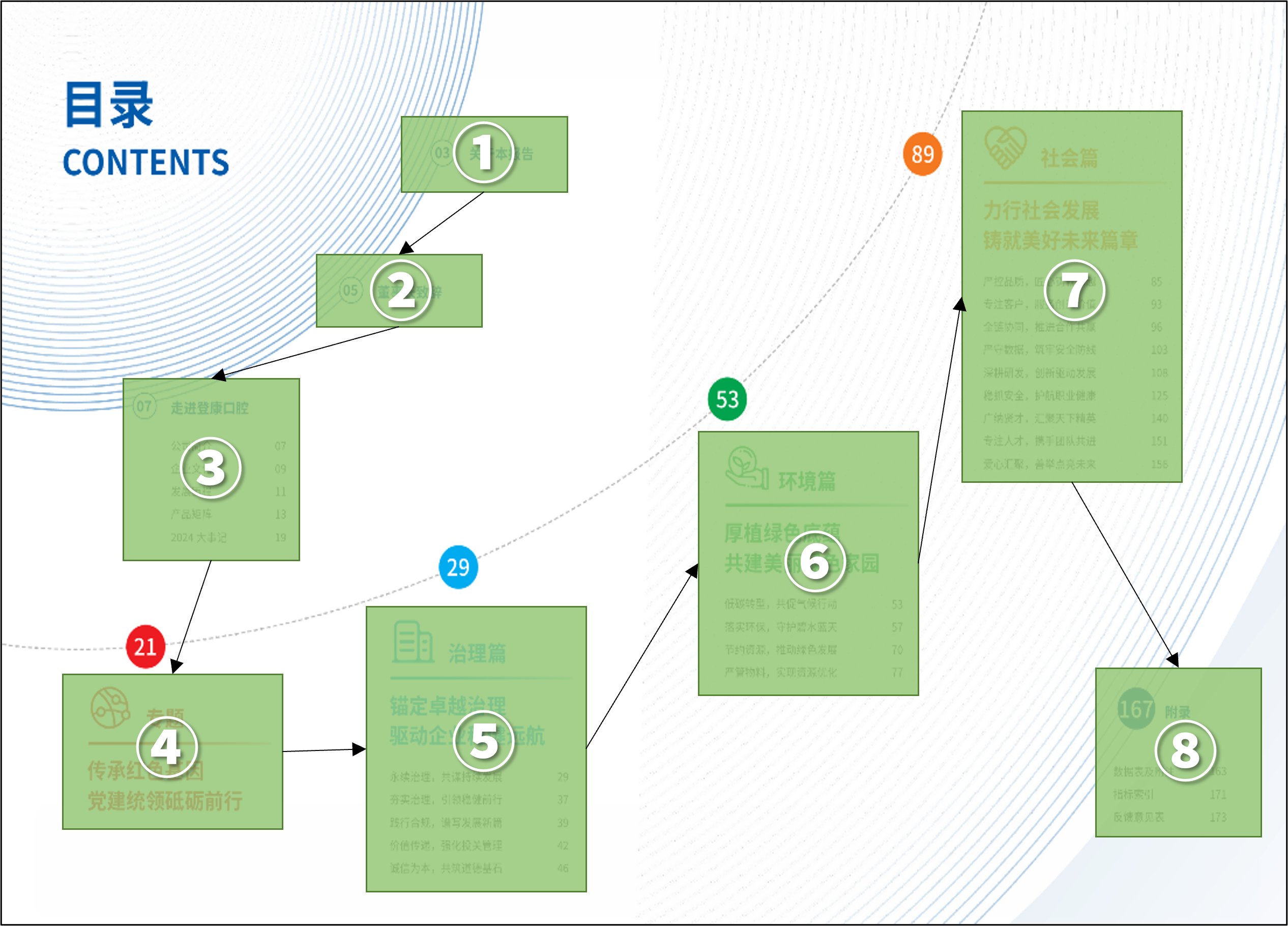}
    \caption{}
    \label{fig:sub1}
  \end{subfigure}
  \hfill
  \begin{subfigure}[b]{0.47\columnwidth}
    \includegraphics[width=\linewidth]{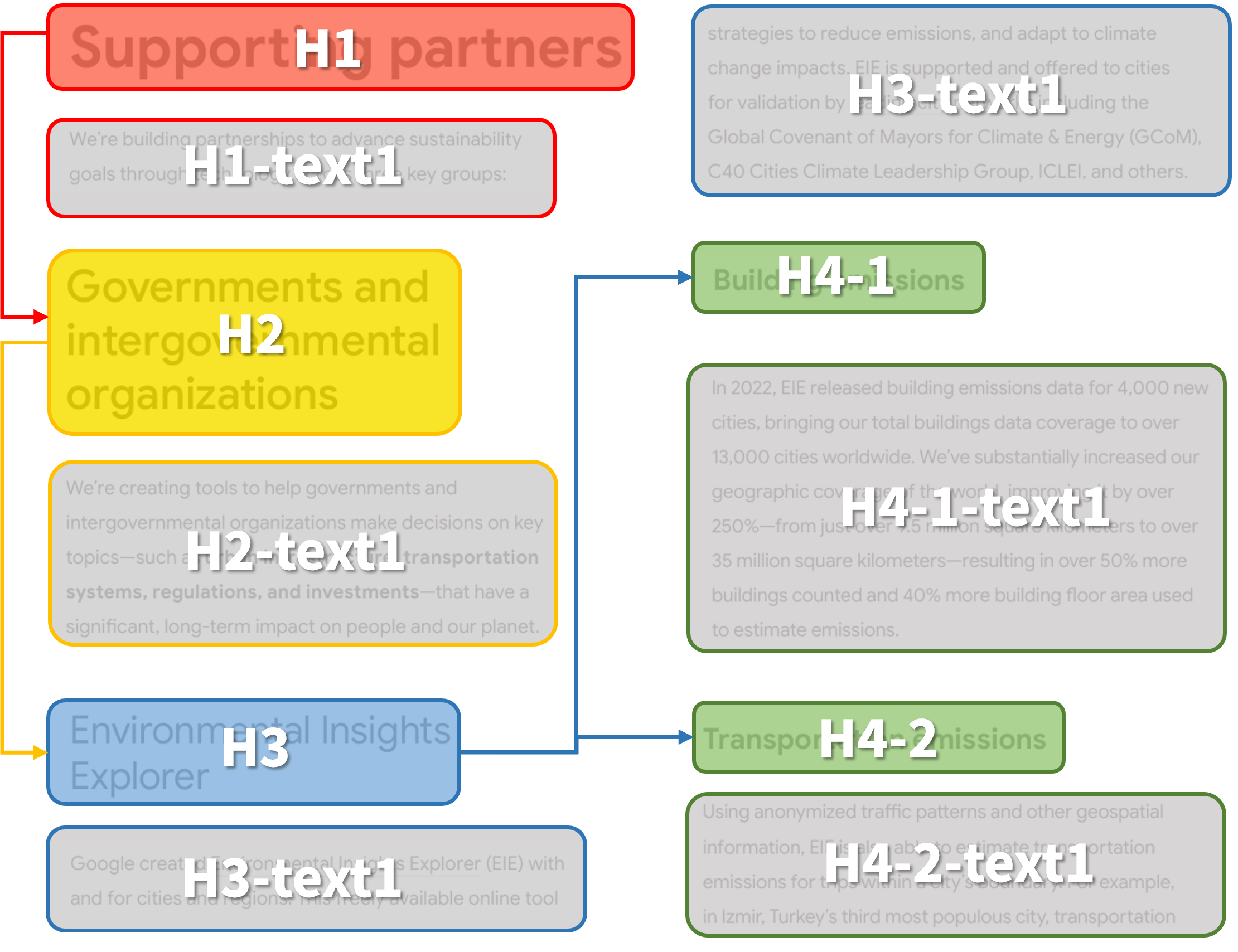}
    \caption{}
    \label{fig:sub2}
  \end{subfigure}

  \caption{Representative Challenges in ESG Report Parsing.}
  \label{fig:Challenges}
\end{figure}


\begin{figure}[t]
  \centering
  \includegraphics[width=\columnwidth]{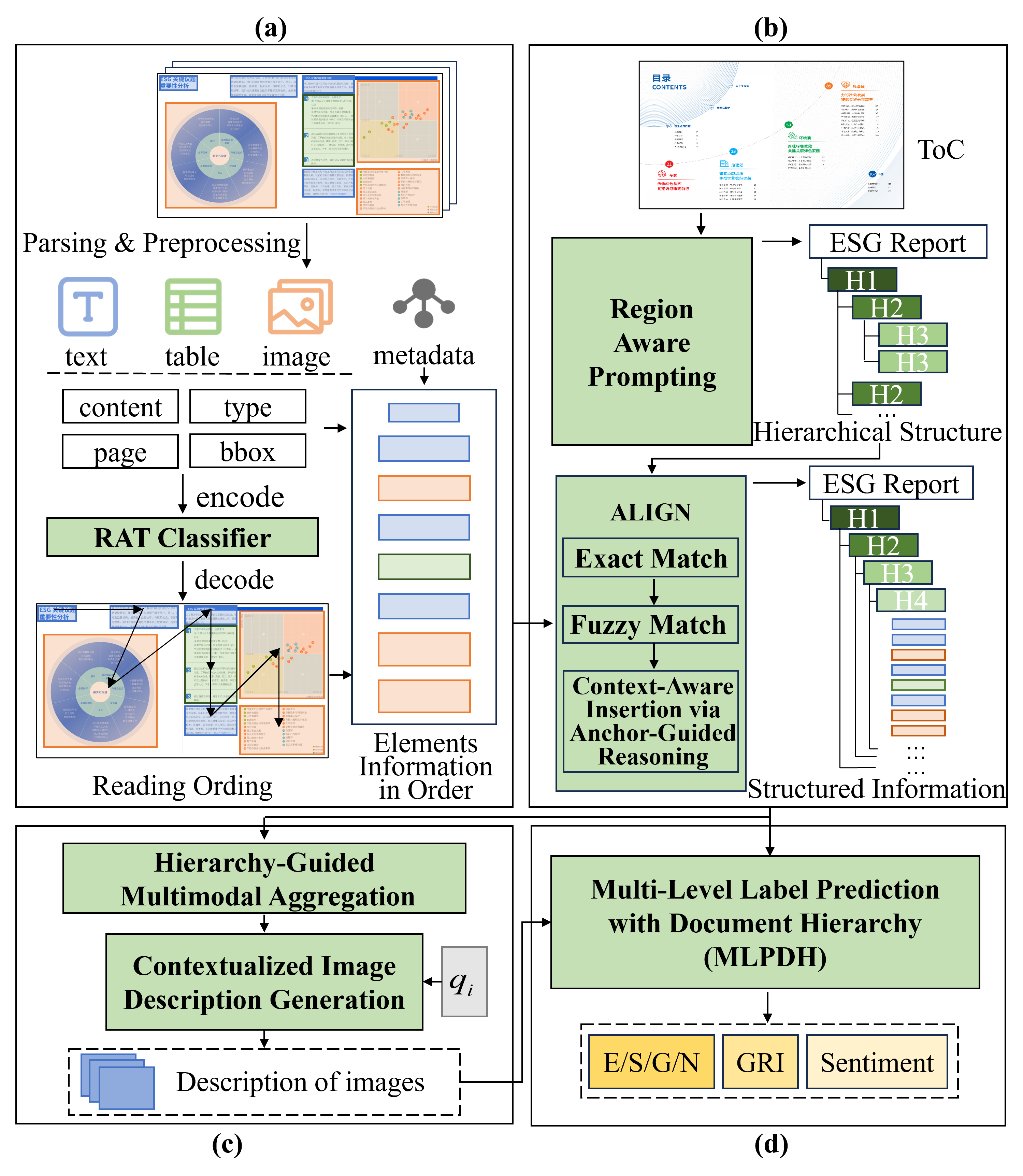}
  \caption{Overview of the Pharos-ESG.}
  \label{fig:overview}
\end{figure}

As global finance shifts toward sustainability, ESG principles are becoming embedded in the institutional fabric of capital allocation and regulatory governance. With regulators worldwide transitioning ESG disclosures from voluntary to mandatory, they now serve as critical infrastructure linking firms, investors, and regulators in the pursuit of long-term value and systemic sustainability.

However, ESG reports, the primary medium for sustainability disclosures, are typically released as visually dense, lengthy PDFs, posing two key technical challenges. First, their visual layout is highly heterogeneous, with interleaved text, tables, and charts organized in complex, often slide-like formats. This inconsistency complicates both layout parsing and reading-order inference, even in ostensibly structured sections like directories (Figure~\ref{fig:sub1}). Second, their content hierarchy is largely implicit. These reports frequently exceed 50 pages and lack standardized structural indicators such as numbered headings or consistent formatting, making it difficult to recover hierarchical organization (Figure~\ref{fig:sub2}).

These challenges have driven financial research to adopt indirect proxies for ESG content, such as simple disclosure indicators~\cite{Ni2019Mandatory}, small-scale case studies~\cite{Berg2022AggregateConfusion}, or third-party ratings~\cite{Christensen2022doclaynet}, all of which bypass the rich semantics of the reports themselves. Meanwhile, existing document parsers, designed for structurally regular formats like academic papers, legal contracts, or forms~\cite{Palm2019Attend, Xu2020layoutlm, Huang2022LayoutLMv3}, struggle with the irregularity and implicit structure of ESG reports, limiting their applicability to this domain.

To address these challenges, we present \textbf{Pharos-ESG}, a unified framework for multimodal parsing, contextual grounding, and hierarchical labeling of ESG reports (Figure~\ref{fig:overview}). Pharos-ESG integrates four core components: (a) reading-order modeling based on layout flow, (b) structure reconstruction guided by table-of-contents (ToC) anchors, (c) contextual transformation of visual elements into natural language, and (d) multi-level labeling across ESG topics, GRI indicators, and sentiment. The system outputs reading-order-aligned structures with linked text, tables, and images, enriched by hierarchical and semantic annotations. Experiments on expert-annotated benchmarks demonstrate that Pharos-ESG consistently outperforms both specialized parsers (e.g., MinerU~\cite{Huang2022LayoutLMv3}, Marker~\cite{Tito2023Hierarchical}, Docling~\cite{Kim2023Donut}) and general-purpose multimodal models (e.g., GPT-4o, Gemini 2.5 Pro, Doubao-Seed) across multiple tasks.

To facilitate real-world ESG analysis and document understanding, we release \textbf{Aurora-ESG} (\textbf{A}nnotated \textbf{U}nified \textbf{R}epository of \textbf{O}mnidimensional \textbf{R}eport \textbf{A}nalytics for ESG disclosures) based on Pharos-ESG outputs.
As the first large-scale public ESG report dataset, Aurora-ESG includes over 24K disclosures from Mainland China, Hong Kong, and U.S. markets, totaling more than 8 million content blocks.

\paragraph{\textbf{Contributions}}
(1) We propose Pharos-ESG, enabling coherent interpretation of ultra-long, complex ESG reports for accurate and structured understanding of ESG disclosures.
(2) We evaluate Pharos-ESG across diverse markets, confirming its robustness to multilingual and structurally diverse ESG disclosures in real-world settings.
(3) We release Aurora-ESG, addressing the lack of high-quality resources for multimodal and hierarchical understanding of real-world ESG reports in financial contexts. (4) We open-source our code, models, and a subset of the Aurora-ESG dataset to support reproducibility: \url{https://github.com/liucun-zy/Pharos-ESG}. Due to the large scale of the dataset, the full version is available upon request from the authors for research purposes.

\section{Related Work}
Recent advances in document intelligence have significantly improved performance on general-purpose document understanding tasks~\cite{Xu2020layoutlm, Huang2022LayoutLMv3}. However, ESG reports remain challenging due to chaotic reading order from irregular, slide-like layouts and implicit hierarchies caused by lengthy, weakly structured content.

\paragraph{Parsing Systems and VLMs}  Dedicated document parsing systems such as Docling~\cite{livathinos2025docling}, MinerU~\cite{wang2024mineru}, Marker~\cite{marker2025}, and TextIn~\cite{textin2023}, are effective at converting documents into structured outputs. They often incorporate state-of-the-art layout analysis models like EfficientViT~\cite{liu2023efficientvit}, RT-DETR~\cite{zhao2024detrs}, and LayoutLMv3~\cite{Huang2022LayoutLMv3}, achieving high accuracy on datasets with regular layouts such as PubLayNet~\cite{zhong2019publaynet}. 
However, their performance degrades significantly on structurally diverse datasets like DocLayNet~\cite{pfitzmann2022doclaynet}, revealing limitations in generalization to visually irregular documents, including ESG reports.
In parallel, vision-language models (VLMs) such as GPT (OpenAI), Gemini (Google), and DeepSeek (DeepSeek) provide a flexible, end-to-end solution by directly converting page images into text. However, these models often hallucinate and incur substantial computational costs when reconstructing implicit hierarchies in long, weakly structured ESG reports.

\paragraph{Cross-Modal Semantic Alignment} 
Transformer-based models such as DocFormer~\cite{Appalaraju2021DocFormer}, LayoutLMv3~\cite{Huang2022LayoutLMv3}, StrucTexT~\cite{Appalaraju2021DocFormer}, UDoc~\cite{Kim2023Donut}, and SelfDoc~\cite{Li2021SelfDoc} employ cross-modal attention to associate figures with captions and link textual blocks with related tables or legends. While effective on structurally regular documents, they struggle with ESG reports, where semantically related elements are often dispersed across non-adjacent, even cross-page regions.

\paragraph{Document Hierarchical Structure Modeling}
Traditional approaches typically model structure at the page level, limiting global context and failing to capture cross-page semantic and hierarchical links~\cite{Huang2022LayoutLMv3, Xia2022Structured}. Most rely on explicit visual cues (e.g., numbering, indentation, font size) which often prove unreliable in complex or inconsistent layouts~\cite{Wang2023DocParser, zhong2019publaynet, Xu2020layoutlm}. Recent models combine multimodal encoders with structure-aware decoders for joint hierarchy prediction, such as HRDoc~\cite{Marulanda2023Correspondence} and DINO~\cite{Zhang2023DINO}. Yet, attention decay in lengthy documents (e.g., 20+ pages) still leads to recognition failures~\cite{Beltagy2020Longformer, Huang2022LayoutLMv3}. These challenges are amplified in ESG reports, which feature exceptional length, weak structural cues, and diverse formatting across issuers.

\paragraph{ESG Data Resources} 
To our knowledge, four recent datasets support ESG-related research. \citet{morio2023nlp} released 10{,}000+ corporate climate policy documents. ESG-FTSE~\cite{wang2024esg} contains 3{,}913 news articles on the UK’s top 10 firms. A3CG~\cite{ong2025towards} compiles ESG reports from 1{,}679 Singapore-listed companies, and DynamicESG~\cite{tseng2023dynamicesg} includes 2{,}220 Taiwan-based ESG news articles. However, these datasets offer limited coverage of full-length, multimodal ESG reports and lack the fine-grained annotations required for downstream financial tasks.

\section{Methodology}
This section outlines the proposed Pharos-ESG system, with its overall architecture depicted in Figure~\ref{fig:overview}.

\subsection{Reading Order Modeling}

Based on the extraction of multimodal elements from ESG reports via a structured preprocessing pipeline---including metadata extraction~\cite{han2003automatic}, layout analysis~\cite{Huang2022LayoutLMv3}, content parsing~\cite{zhang2020trie}, and noise reduction~\cite{easyocr2024}---Pharos-ESG adopts and extends the Immediate Succession During Reading paradigm~\cite{zhang2024modeling} for reading order modeling, transitioning from global sequence ranking to a pairwise succession classification framework.

\paragraph{\textbf{Block Content Encoding}}  
Each page is represented as a set of content blocks: 
\begin{equation}
\mathcal{D}_p = \{(w_i, b_i, c_i, p)\}_{i=1}^{N_p},
\end{equation}
where \(w_i\) is the block content; \(b_i\) is the bounding box; \(c_i\) is the block type; and \(p\) is the page index. \(N_p\) is the number of blocks on page \(p\).

\paragraph{\textbf{Multimodal Feature Construction}}
For each ordered block pair $(i, j)$, we construct a feature vector $\varphi_{ij}$ by integrating semantic, spatial, and categorical signals:
\begin{equation}
\varphi_{ij} =
\begin{bmatrix}
E(w_i),\ E(w_j),\ \Delta y_{ij},\ \Delta x_{ij}, \\
\mathrm{IoU}(b_i, b_j),\ \mathrm{Dist}(b_i, b_j),\ E(c_i),\ E(c_j)
\end{bmatrix},
\end{equation}
where $E(w_i)$ and $E(w_j)$ are content embeddings: text and tables  are encoded using LayoutLMv3~\cite{Huang2022LayoutLMv3} For image blocks, the local image URL is used only as a placeholder for block identity and is not encoded nor used as a visual feature in the model. $\Delta y_{ij}$ and $\Delta x_{ij}$ denote center offsets; $\mathrm{IoU}{ij}$ and $\mathrm{Dist}{ij}$ are the bounding-box overlap and normalized distance; and $E(c_i), E(c_j)$ are one-hot encodings of block types.

\paragraph{\textbf{Relation Prediction Module}}  
A Relation-Aware Transformer (RAT) predicts whether block $j$ directly follows block $i$. The feature vector $\varphi_{ij}$ is enhanced via cross-attention between $E(w_i)$ and $E(w_j)$, and passed through a Transformer encoder to compute the succession score:
\begin{equation}
s_{ij} = \sigma(\mathbf{W} \cdot \mathrm{Transformer}(\varphi_{ij}) + \mathbf{b}),
\end{equation}
where $s_{ij} \in [0, 1]$ is the predicted probability, and $\mathbf{W}, \mathbf{b}$ are learnable parameters.

\paragraph{\textbf{Reading Order Label Generation}}  
A directed edge $(i \rightarrow j)$ is created if $s_{ij} > \tau$, where $\tau$ is a tunable threshold empirically set in the range $[0.2, 0.5]$ to accommodate layout variability in ESG reports. Since multiple successors may satisfy this condition for a given block $i$, a directed weighted graph is constructed with $s_{ij}$ as edge weights. The final reading sequence is then obtained via topological sorting, ensuring a globally consistent, acyclic order.

\subsection{ToC-guided hierarchical structure reconstruction} 

To capture implicit hierarchies in ESG reports, Pharos-ESG introduces a ToC-centered framework comprising: (1) a ToC parser that extracts section-level cues, and (2) an alignment module that enforces structural consistency between the ToC and the document body.

\paragraph{\textbf{ToC Structure Parsing}}
To parse visually diverse ToC layouts in ESG reports, we propose \textbf{RAP} (\textbf{R}egion-\textbf{A}ware \textbf{P}rompting), a visual prompting strategy that leverages color similarity, spatial proximity, and textual adjacency to guide multimodal large language models (MLLMs) in modeling ToC entries and their context for implicit hierarchy inference. RAP consists of four components: cross-region entry aggregation, context-aware label enrichment, region-based hierarchy inference, and multi-line consolidation.

\paragraph{\textbf{ToC-Body Alignment with Anchor-Guided Reasoning}}
To align the hierarchical structure parsed from ToC with the corresponding content in the document body, we propose \textbf{ALIGN} (\textbf{A}nchor-based \textbf{L}inguistic \textbf{I}ndexing for \textbf{G}ranular \textbf{N}avigation), a multi-stage alignment algorithm designed for visually heterogeneous layouts. ALIGN operates in three stages: (1) \textit{exact match}, performing character-level matching to identify high-confidence anchors; (2) \textit{fuzzy/containment matching}, using Levenshtein similarity~\cite{Mehlhorn1990Dynamic} and substring containment to expand coverage; and (3) \textit{context-aware insertion}, resolving unmatched ToC headings by reasoning within anchor-defined windows in the document body. To this end, a large language model is prompted using a tailored \textbf{CIP} (\textbf{C}ontext-aware \textbf{I}nsertion \textbf{P}rompt) to perform four structured inference steps: (i) summarizing paragraphs to generate semantic representations; (ii) assessing whether the region lacks an overview heading; (iii) identifying where the unmatched heading improves local structure; and (iv) selecting the optimal insertion point based on structural and semantic alignment.

\subsection{Image-to-text Transformation}
Pharos-ESG converts visual elements into structured text via a two-stage pipeline informed by context and reading order.

\paragraph{Hierarchy-Guided Multimodal Aggregation}
This component integrates each target image with its surrounding content into a coherent multimodal cluster, guided by nearby section headings and preserving reading order to ensure semantic continuity and structural integrity.

\paragraph{Contextualized Image Description Generation} 
This component transforms visual data into structured language via a two-stage process.
In the \textit{structured semantic modeling} stage, each multimodal cluster is encoded as:
\begin{equation}
    I_i = (h_i, \{x_1, x_2, \ldots, x_k\}, q_i),
\end{equation}
where $h_i$ is the hierarchical heading path, $\{x_1, \ldots, x_k\}$ are the ordered elements, and $q_i$ is the task instruction. The prompt comprises (i) heading path embedding, linking the cluster to its section context, and (ii) element declaration templates, specifying type, order, and content.
In the \textit{multimodal embedding and semantic generation} stage, visual features are extracted by ViT~\cite{Dosovitskiy2020Animage}, projected into a unified semantic space via Q-Former~\cite{Li2023BLIP-2}, and embedded into prompts with structural tags. The full sequence is then processed by Qwen2.5-VL-Instruct~\cite{Bai2023Qwen-vl} to generate descriptions. A case example is shown in Figure~\ref{fig:case_study} (see Section~\nameref{sec:case_study}).

\subsection{Generation of Multidimensional Financial Market Labels}

To support fine-grained analysis in financial applications, we propose \textbf{MLPDH} (\textbf{M}ulti-\textbf{L}evel \textbf{P}rediction with \textbf{D}ocument \textbf{H}ierarchy), a hierarchical classification framework for multilayer label prediction in ESG disclosures. Each content block is annotated with:
(1) a ESG-N category; 
(2) a GRI indicator; 
(3) a sentiment label.
MLPDH follows a three-stage pipeline: 
\emph{ternary embedding} $\rightarrow$ \emph{hierarchical attention} $\rightarrow$ \emph{hierarchy-aware prediction}.

\paragraph{Ternary Embedding}
Each content block is represented by a composite embedding that integrates textual semantics, hierarchical context, and global reading order:
\begin{equation}
    \mathbf{e}_{\text{blk}} = \mathbf{E}_{\text{text}} + \mathbf{E}_{\text{lvl}} + \mathbf{E}_{\text{pos}},
\end{equation}
where $\mathbf{E}_{\text{text}}$ is obtained from the \texttt{[CLS]} token of Chinese-RoBERTa-wwm-ext~\cite{Sun2021ChineseBERT}; $\mathbf{E}_{\text{lvl}}$ encodes the heading path $\{h_1, h_2, h_3, h_4\}$ via a GRU:
\begin{equation}
    \mathbf{E}_{\text{lvl}} = \mathbf{W}_{\text{lvl}} \cdot \texttt{GRU}([\text{Emb}(h_1), \ldots, \text{Emb}(h_4)]);
\end{equation}
and $\mathbf{E}_{\text{pos}}$ captures the block’s global reading order.

\paragraph{Hierarchical Attention}
To extract level semantics, we apply stacked attention to propagate hierarchical signals. For each level 
$h$, the semantic vector is:
\begin{equation}
    \mathbf{v}^{(h)}_{\text{blk}} = \text{softmax} \left( 
    \tfrac{ 
        (\mathbf{W}_q \mathbf{e}_{\text{blk}})^\top 
        (\mathbf{W}_k \mathbf{v}^{(h-1)}_{\text{ref}}) 
    }{ \sqrt{d} } 
    \right) \cdot \mathbf{W}_v \mathbf{v}^{(h-1)}_{\text{ref}}
\end{equation}
with $\mathbf{v}^{(0)}_{\text{ref}} = \mathbf{E}_{\text{lvl}}$. 

\paragraph{Hierarchy-Aware Prediction}
Each level's label is predicted via sigmoid classification, with hierarchical consistency enforced by a parent-child constraint that penalizes violations of label dependencies:
\begin{equation}
    \mathcal{L}_{\text{hier}} = \sum_{h=2}^H \sum_{c^h} \max\left(0, P(c^h) - P(\text{parent}(c^h))\right),
\end{equation}
where $P(c^h)$ is the predicted probability for label $c^h$ at level $h$. The final objective combines binary cross-entropy loss with the hierarchical constraint:
\begin{equation}
    \mathcal{L}_{\text{total}} = \sum_{h=1}^H \text{BCE}(P^h, Y^h) + \lambda \cdot \mathcal{L}_{\text{hier}}.
\end{equation}

During inference, labels with $P > \theta$ (default: 0.5) are selected to form a coherent multi-level label path (e.g., \texttt{E}~$\rightarrow$~\texttt{e-gri30}~$\rightarrow$~\texttt{negative}).

\section{Experiment}

\begin{table*}[!t]
\centering
\scriptsize
\setlength{\tabcolsep}{0.5pt}
\renewcommand{\arraystretch}{1}

\begin{tabular}{
>{\centering\arraybackslash}p{1.7cm}  
>{\centering\arraybackslash}p{2.3cm}
*{4}{>{\centering\arraybackslash}p{1.05cm}}
*{6}{>{\centering\arraybackslash}p{1.05cm}}
>{\centering\arraybackslash}p{2.5cm}
}
\hline
\multirow{3}{*}{\textbf{Category}}
& \multirow{3}{*}{\textbf{Method}} 
& \multicolumn{4}{c}{\textbf{Comprehensive ESG Report Analysis}} 
& \multicolumn{6}{c}{\textbf{ToC Extraction}} 
& \textbf{Hierarchy Alignment} \\

\cline{3-13}

& & \multirow{2}{*}{\textbf{Prec.}} & \multirow{2}{*}{\textbf{Rec.}} & \multirow{2}{*}{\textbf{F1}} & \multirow{2}{*}{\textbf{ROKT}}
& \multicolumn{2}{c}{\textbf{CC (\%)}} 
& \multicolumn{2}{c}{\textbf{RC (\%)}} 
& \multicolumn{2}{c}{\textbf{HC (\%)}} 
& \multirow{2}{*}{\textbf{TBTA (\%)}} \\

\cline{7-12}
& & & & & 
& \textbf{GP} & \textbf{RAP} 
& \textbf{GP} & \textbf{RAP} 
& \textbf{GP} & \textbf{RAP} & \\

\hline
\multirow{4}{*}{\begin{tabular}[c]{@{}c@{}}Dedicated\\Document\\Parsers\end{tabular}}
& Marker & 41.74 & 29.53 & 35.48 & 0.22 & - & - & - & - & - & - & \multicolumn{1}{c}{2.87} \\
& Docling & 72.07 & 75.31 & 72.61 & 0.59 & - & - & - & - & - & - & \multicolumn{1}{c}{14.35} \\
& MinerU & 73.46 & 77.56 & 72.41 & 0.62 & - & - & - & - & - & - & \multicolumn{1}{c}{4.73} \\
& Textin & 87.43 & 74.21 & 80.14 & 0.60 & - & - & - & - & - & - & \multicolumn{1}{c}{7.48} \\
\hline
\multirow{6}{*}{\begin{tabular}[c]{@{}c@{}}General-purpose\\Multimodal\\Models\end{tabular}}
& DeepSeek-V3 & 40.11 & 57.45 & 47.28 & 0.38 & 62.35 & 82.74 & 54.12 & 87.21 & 67.31 & 90.81 & \multicolumn{1}{c}{33.27} \\
& Qwen3 & 45.83 & 53.20 & 48.26 & 0.58 & 92.25 & \textbf{100} & \textbf{94.34} & 97.00 & \textbf{90.28} & 96.24 & \multicolumn{1}{c}{54.43} \\
& DeepSeek-R1 & 65.24 & 58.42 & 61.64 & 0.54 & 80.82 & 91.00 & 85.32 & 92.64 & 75.21 & 96.35 & \multicolumn{1}{c}{53.68} \\
& Doubao-Seed & 62.51 & 62.45 & 63.73 & 0.35 & \textbf{93.14} & \textbf{100} & 79.28 & 95.00 & 87.37 & 95.20 & \multicolumn{1}{c}{40.57} \\
& GPT-4o & 63.90 & 62.83 & 63.29 & 0.71 & 83.05 & 97.00 & 86.25 & \textbf{100} & 69.13 & \textbf{97.00} & \multicolumn{1}{c}{41.35} \\
& Gemini 2.5 Pro & 84.45 & 86.78 & 86.53 & 0.72 & 92.01 & 95.32 & 91.27 & 94.54 & 80.09 & 91.35 & \multicolumn{1}{c}{63.30} \\
\hline
\textbf{Ours}& \textbf{Pharos-ESG} & \textbf{91.23} & \textbf{93.00} & \textbf{92.57} & \textbf{0.90} & 79.62 & 91.37 & 77.67 & 95.52 & 67.31 & 92.71 & \multicolumn{1}{c}{\textbf{91.78}} \\
\hline
\end{tabular}
\caption{Experimental results of different methods. (Prec.: precision, Rec.: recall)}
\label{tab:overall}
\end{table*}

In this section, we conduct some extensive experiments to evaluate the proposed framework, Pharos-ESG.

\subsection{Data}
    To evaluate the parsing capabilities of Pharos-ESG, we focus on Chinese ESG reports, which present more complex layouts and structures. We collected 50 reports (2,383 pages) from Wind. To assess Pharos-ESG’s performance in other markets, we additionally sampled 10 reports each from the Hong Kong and U.S. markets (903 pages in total), covering diverse formats and languages. To construct the Aurora-ESG dataset, we ultimately gathered 24,409 reports across all three markets.

\subsection{Experimental Design}

We evaluate Pharos-ESG using expert annotations on 70 ESG reports from three markets, labeled at the document, page, and block levels by three domain experts. After consolidation, annotations were standardized into JSON format as gold-standard references. Based on this, we define baselines and task-specific metrics.

\paragraph{Baselines}
Pharos-ESG is compared to three system groups:
(1) \textit{Dedicated document parsers:} MinerU~\cite{Huang2022LayoutLMv3}, Marker~\cite{Tito2023Hierarchical}, Docling~\cite{Kim2023Donut}, and Textin~\cite{Appalaraju2021DocFormer}, evaluated on reading order prediction, hierarchy alignment, and structural parsing.
(2) \textit{General-purpose multimodal models:} GPT-4o, Gemini 2.5 Pro, Doubao-Seed, DeepSeek-V3, DeepSeek-R1, and Qwen 3, evaluated on ToC Extraction and tasks from the first group.
(3) \textit{Traditional and neural classifiers:} SVM+TF-IDF~\cite{Joachims2005Text}, XGBoost~\cite{Chen2016XGBoost}, BERT-base~\cite{Devlin2019BERT}, HAN~\cite{Huang2021Hierarchical}, and HMCN~\cite{Sadat2022Hierarchical}, evaluated on multi-level label prediction.

\paragraph{Metrics}
Task-specific metrics include:

\begin{itemize}
    \item \textit{Structured document transformation:} precision, recall, and F1; accuracy is undefined due to the absence of negative samples.
    
    \item \textit{Reading order prediction:} Evaluated using Reading Order Kendall’s Tau (ROKT), a reading-order consistency metric derived from Kendall’s Tau~\cite{kendall1938new}.
\end{itemize}

Together, the metrics outlined above provide a complementary assessment of the model’s overall performance.

\begin{itemize}
    \item \textit{ToC Structure Extraction}:
    \begin{itemize}
        \item Content Completeness (CC): $1 - \frac{\text{Redundancy} + \text{Missing}}{\text{Total}}$
        \item Region Order Consistency (RC): $\frac{\text{Correct Order}}{\text{Total}}$
        \item Hierarchical Consistency (HC): $\frac{\text{Correct Classification}}{\text{Total}}$
    \end{itemize}

    \item \textit{ToC-Body Title Alignment (TBTA)}: $\frac{\text{Correct Titles}}{\text{Reference Titles}}$

    \item \textit{Multi-level Label Prediction}: Multi-level F1, macro-F1, and a specialized parent-child consistency metric, Hierarchy Logic Accuracy (HLA): $\frac{\text{Valid Parent-Child Predictions}}{\text{Total Relations}}$

\end{itemize}

\subsection{Overall Performance}

Leveraging long-document support, Pharos-ESG and dedicated systems are evaluated in batch mode on 50 full Chinese reports. In contrast, due to context limits, multimodal models are tested incrementally on 10 reports split into 5-page segments. Results are shown in Table~\ref{tab:overall}.

\paragraph{Comprehensive ESG Report Analysis} Pharos-ESG shows strong overall performance in both structured document transformation and reading order prediction.
For \textit{structured document transformation}, Textin achieves the best performance among dedicated document parsers, with 87.43\% precision, 74.21\% recall, and an F1-score of 80.14\%. In contrast, Gemini 2.5 Pro, representative of general-purpose models, shows greater variability in small-scale settings, reaching an F1 of 86.53\%. Pharos-ESG surpasses all baselines, achieving 91.23\% precision, 93.00\% recall, and 92.57\% F1, demonstrating consistent superiority on ESG-specific parsing.
For \textit{reading order prediction}, general-purpose models, however, suffer from context window constraints and hallucinations, often generating incoherent sequences and lower ROKT. Among dedicated document parsing systems, we observe a positive correlation between ROKT and parsing metrics, suggesting that accurate reading order modeling benefits structural extraction.
Pharos-ESG achieves a ROKT of 0.90, effectively modeling long-range dependencies and maintaining sequence alignment under complex layouts.

\paragraph{ToC Extraction} For general-purpose multimodal models, RAP consistently outperforms general prompting (GP), yielding average improvements of +10.41\% in CC, +12.64\% in RC, and +16.26\% in HC. Several models, including GPT-4o, Doubao-Seed, and Qwen3, even achieve perfect (100\%) scores.
These results highlight RAP's strong robustness and adaptability across model families. To reduce the cost and privacy concerns associated with API-based deployment, Pharos-ESG integrates Qwen2.5-VL-7B-Instruct locally. Even in this constrained setting, RAP maintains competitive performance.

\paragraph{ToC-Body Title Hierarchy Alignment}
Dedicated parsers perform poorly on this task (TBTA $<$ 20\%), as they rely on heuristic features (e.g., font size, indentation) that fail under the diverse and implicit layouts of ESG reports. General-purpose multimodal models perform better but suffer from context limitations and input fragmentation, leading to inconsistent hierarchy predictions (TBTA $<$ 65\%).
In contrast, Pharos-ESG achieves 91.78\% TBTA by leveraging ToC structures as alignment anchors. The ALIGN strategy enables robust end-to-end alignment through multi-stage matching and contextual inference, effectively reconstructing body hierarchies even under complex layouts.

\subsection{Ablation Study}

\begin{table}[t]
\centering
\scriptsize
\setlength{\tabcolsep}{0.5pt}
\renewcommand{\arraystretch}{1.1}
\begin{tabular}{
>{\centering\arraybackslash}m{1cm}  
>{\centering\arraybackslash}m{1.38cm}  
>{\centering\arraybackslash}m{1.38cm}  
>{\centering\arraybackslash}m{1.38cm}  
>{\centering\arraybackslash}m{1cm}  
>{\centering\arraybackslash}m{1cm}  
>{\centering\arraybackslash}m{0.8cm}  
}
\hline
\multirow{2}{*}{\shortstack[c]{\textbf{Config} \\[-2pt] \textbf{ID}}} &
\multirow{2}{*}{\shortstack[c]{\textbf{Reading} \\[-2pt] \textbf{Order} \\[-2pt] \textbf{Modeling}}} &
\multirow{2}{*}{\shortstack[c]{\textbf{ToC} \\[-2pt] \textbf{Structure} \\[-2pt] \textbf{Parsing}}}
&
\multirow{2}{*}{\shortstack{\textbf{ToC-Body}\\\textbf{Alignment}}} & 
\multicolumn{3}{c}{\textbf{Pharos-ESG Performance}} \\
\cline{5-7}
& & & & \textbf{Prec.} & \textbf{Rec.} & \textbf{F1} \\
\hline
1 & \xmark & \xmark & \xmark & 75.10 & 78.90 & 76.95 \\
2 & \xmark & \textit{GP} & \xmark & 78.63 & 79.20 & 78.79 \\
3 & \xmark & \textit{RAP} & \xmark & 84.43 & 82.76 & 83.57 \\
4 & \cmark & \textit{RAP} & \xmark & 86.01 & 89.40 & 88.14 \\
5 & \cmark & \textit{RAP} & EM-FCM & 89.12 & 92.85 & 90.05 \\
6 & \textbf{\cmark} & \textbf{\textit{RAP}} & \textbf{\cmark} & \textbf{92.23} & \textbf{95.00} & \textbf{93.59} \\
\hline
\end{tabular}
\caption{Ablation Study of Pharos-ESG.}
\label{tab:ablation}
\end{table}

We performed ablation studies (Table~\ref{tab:ablation}) by incrementally enabling three core components---reading order modeling, ToC structure parsing, and ToC-body alignment---to assess their impact on Pharos-ESG’s overall performance.

Starting from the baseline (Config 1), where all core modules are disabled, the system achieves an F1 score of 76.95, comparable to general-purpose document parsers. Enabling the GP-based ToC parser (Config 2) improves precision from 75.10 to 78.63 by correcting heading levels, though its handling of irregular section labels remains limited. Incorporating the RAP strategy in Config 3 further boosts precision to 84.43, leveraging visual and semantic cues for more accurate hierarchy inference.
Adding the reading order module in Config 4 brings the F1 score to 88.14, with notable improvements in recall due to better modeling of long-range dependencies. 
Config 5 integrates the ToC-body alignment module using only the first two steps of ALIGN, namely Exact Match and Fuzzy/Containment Matching (EM-FCM), resulting in a 3.45-point increase in recall.
Finally, Config 6, the complete Pharos-ESG system, achieves the highest performance, showcasing the synergistic benefits in managing complex ESG layouts.

\subsection{Multi-Level Label Prediction in Financial Markets}

\begin{table}[t]
\scriptsize
\setlength{\tabcolsep}{7pt}
\renewcommand{\arraystretch}{1}
\centering

\begin{tabular}{lccccc}
\hline
\multirow{2}{*}{\textbf{Method}} & \multicolumn{3}{c}{\textbf{Multi-level F1-score}} & \multirow{2}{*}{\textbf{Macro-F1}} &\multirow{2}{*}{\textbf{HLA}} \\
\cline{2-4}
 & \textbf{ESGN} & \textbf{GRI} & \textbf{Sentiment} & \\
\hline

SVM+TF-IDF & 72.14 & 61.59 & 68.31 & 67.35 &  - \\
XGBoost & 75.33 & 65.21 & 71.18 & 70.57 & - \\
BERT-base & 80.21 & 72.30 & 77.61 & 76.71 & 81.31 \\
HAN & 81.51 & 74.11 & 78.93 & 78.18 & 82.12 \\
HMCN & 82.70 & 76.86 & 79.07 & 79.54 & 88.15 \\
\hline
\textbf{MLPDH} & \textbf{85.62} & \textbf{84.23} & \textbf{89.11} & \textbf{86.32} & \textbf{94.78} \\
\hline
\end{tabular}
\caption{Multi-level Label Prediction Performance.}
\label{tab:multi_level_f1_scores}
\end{table}

Based on structured ESG report data, this section evaluates the MLPDH module within Pharos-ESG, which maps content blocks to a three-level label structure: ESGN category $\rightarrow$ GRI indicator $\rightarrow$ sentiment. The evaluation is conducted on 15{,}213 expert-annotated blocks (10{,}213 train / 1{,}500 validation / 3{,}000 test) collected from 50 ESG reports.

As shown in Table~\ref{tab:multi_level_f1_scores}, MLPDH significantly outperforms all baselines. It achieves a macro-F1 score of 86.32, surpassing the strongest baseline (HMCN) by 6.78\%, with the most notable improvements observed at the sentiment level. Most baseline models obtain multi-level F1 scores below 80, with SVM+TF-IDF and XGBoost performing around 70 due to their limited semantic modeling capabilities.

In HLA, MLPDH scores highest at 94.78. BERT-base drops to 81.31 due to parent-child inconsistencies from missing cross-level constraints. HAN, though using hierarchical attention, trails MLPDH by 12.66\%. HMCN adds hierarchical prediction but lacks triplet embedding and cross-level attention, causing a 6.63\% drop.

\subsection{Cross-Market Generalization of Pharos-ESG}

\begin{table}[t]
\centering
\scriptsize
\setlength{\tabcolsep}{5pt}
\renewcommand{\arraystretch}{1}

\begin{tabular}{lcccc}
\hline
\multirow{2}{*}{\textbf{Market}} & \multicolumn{3}{c}{\textbf{Comprehensive ESG Report Analysis}} & \textbf{Label Prediction}\\
\cline{2-5}
&
\textbf{Parsing F1} & \textbf{ROKT} & \textbf{TBTA} & \textbf{Macro-F1} \\
\hline
China stock & 92.23 & 0.92 & 92.46 & 86.32 \\
HK stock & 89.05 & 0.88 & 89.50 & 87.20 \\
US stock & 94.30 & 0.94 & 94.80 & 87.60 \\
\hline
\end{tabular}
\caption{Cross-Market Performance of Pharos-ESG.}
\label{tab:cross_performance}
\end{table}

Pharos-ESG is tested on Hong Kong and U.S. reports to assess its adaptability across languages, formats, and document structures. Table~\ref{tab:cross_performance} shows results on comprehensive ESG report analysis and financial label prediction.

\paragraph{Comprehensive ESG Report Analysis} On Hong Kong reports, Pharos-ESG shows slightly lower Parsing F1, ROKT, and TBTA than on China reports. In contrast, it performs better on U.S. reports, likely due to more standardized formats, clearer hierarchies, and consistent block segmentation---factors that aid structure extraction and reading order modeling.

\paragraph{Financial Label Prediction} Pharos-ESG performs better on Hong Kong and U.S. markets than on China data. The Hong Kong set yields a macro-F1 of 87.20, likely due to more consistent structure and clearer terminology. The U.S. market performs best, achieving 87.60 macro-F1, benefiting from standardized language, well-defined sectioning, and uniform disclosure practices.

Overall, the results confirm Pharos-ESG’s strong generalization across markets and languages, enabling robust ESG report parsing.

\subsection{Case Study}
\label{sec:case_study}

\begin{figure}[t]
  \centering
  \includegraphics[width=\columnwidth]{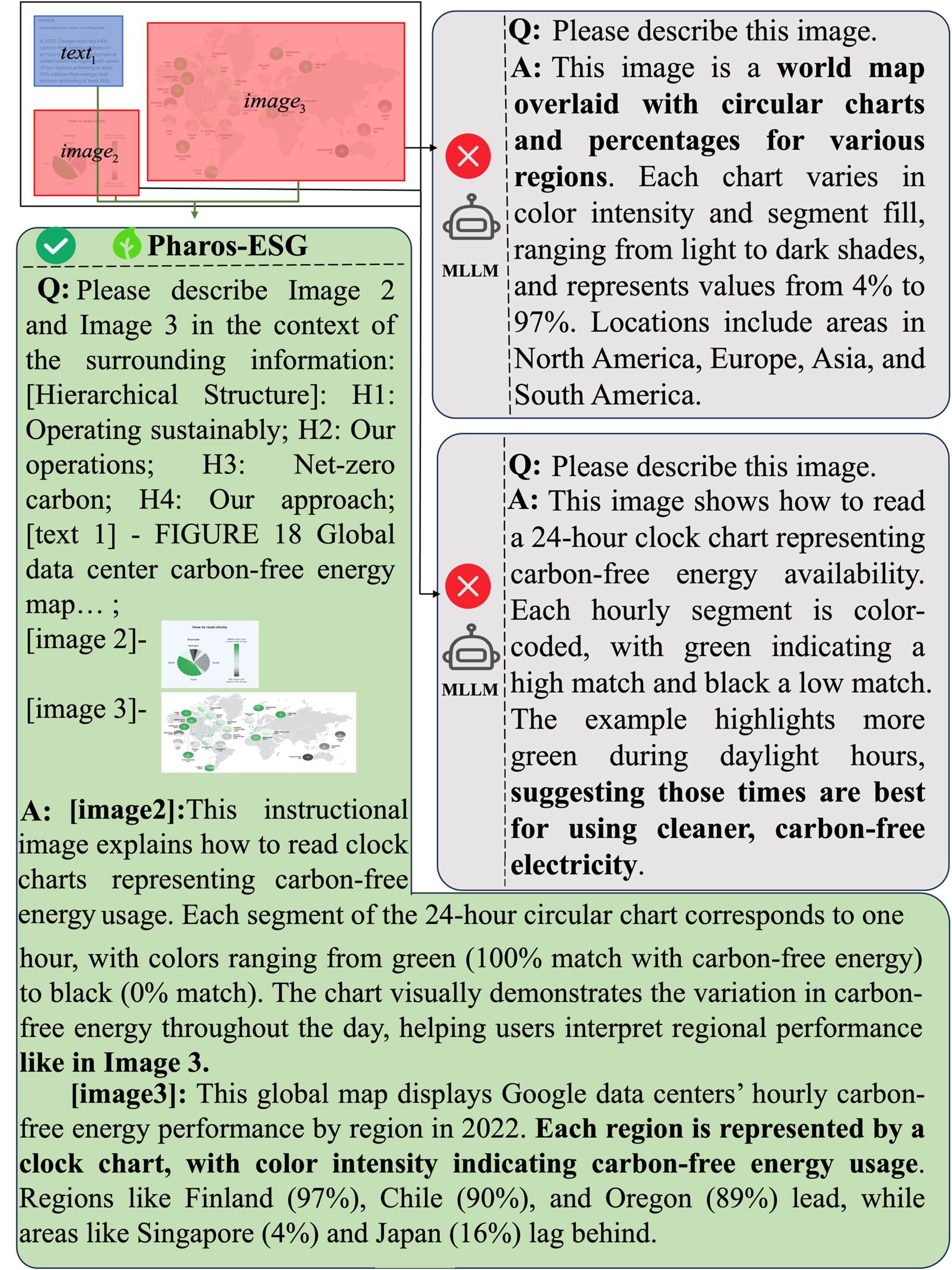}
  \caption{Case studies showing Pharos-ESG’s advantages in semantic understanding and reasoning.}
  \label{fig:case_study}
\end{figure}

\begin{table*}[t]
\centering
\scriptsize
\setlength{\tabcolsep}{1pt}
\renewcommand{\arraystretch}{1}

\begin{tabular}{
  >{\centering\arraybackslash}p{3cm}
  >{\centering\arraybackslash}p{2.7cm}
  >{\centering\arraybackslash}p{4cm}
  >{\centering\arraybackslash}p{4cm}
  >{\centering\arraybackslash}p{3.5cm}
}
\hline
\textbf{Level} & \textbf{Category} & \textbf{Field} & \textbf{Description} & \textbf{Example} \\
\hline

\multirow{8}{*}{\textbf{Document Level}}
& \multirow{4}{*}{Document Identifiers}
    & stock\_code         & Stock code                                & 300XXX.SZ \\
&   & company\_name       & Company name                              & FxXxxx Pharma \\
&   & report\_year        & Reporting year                               & 2024 \\
&   & report\_title       & Title of the ESG report                   & ESG Report 2024 \\
\cline{2-5}
& \multirow{3}{*}{Source Metadata}
    & report\_type        & Report type                               & ESG Report \\
&   & market              & Market the report belongs to              & China \\
&   & original\_filename  & Original JSON file name                        & 300XXX.SZ-...json \\
\hline
\multirow{4}{*}{\textbf{Page Level}}
& Indexing         
    & page\_idx           & Page index (starting from 1)                 & 1, 2, 3... \\
\cline{2-5}
& \multirow{3}{*}{Rendering Links} 
    & page\_markdown\_url & Markdown image link                         & ![](300XXX...1.jpg) \\
&   & page\_file\_url / page\_relative\_path & Local image file path         & /mnt/data/.../1.jpg \\
&   & page\_http\_url     & Online accessible HTTP path                 & http://.../1.jpg \\

\hline

\multirow{10}{*}{\textbf{Content Block Level}}
& Hierarchical depth
    & h1--h4      & Structural depth of section headings        & Environmental Management \\
\cline{2-5}
& \multirow{2}{*}{Content type and payload}
    & data\_type          & Block type      & text / table / image \\
&   & data              & Block content     & ``...company 2024 energy..." \\
\cline{2-5}
& \multirow{3}{*}{Visual Resource Links} 
    & markdown\_url         & Markdown link (table/image)                 & ![](./temp\_images/xxx.jpg) \\
&   & file\_url / relative\_path & Local high-res path                   & /mnt/data/...jpg \\
&   & http\_url          & Online access path                          & http://.../12.jpg \\
\cline{2-5}
& Ordering           
    & reading\_order      & In-page reading order               & 0, 1, 2... \\
\cline{2-5}
& \multirow{3}{*}{Semantic tag} 
    & esg\_category\_label & ESG category label                         & E / S / G / N \\
&   & gri\_label          & One of the 32 GRI labels                    & Energy \\
&   & sentiment\_label    & Sentiment polarity                          & Positive / Neutral / Negative \\
\hline
\end{tabular}
\caption{Hierarchical Field Structure of Each Structured ESG Report in Aurora-ESG.}
\label{tab:Aurora-ESG2}
\end{table*}

Figure~\ref{fig:case_study} provides a representative case study that underscores the contextual and reasoning capabilities of Pharos-ESG. By analyzing a complex page layout consisting of two distinct images and a textual block, it becomes evident that interpreting these visual elements in isolation leads to a significant loss of their semantic essence, as demonstrated by the baseline MLLM outputs. Conversely, Pharos-ESG effectively models the structural interdependence during the description generation process; specifically, it identifies \textit{image\textsubscript{2}} as the requisite legend for \textit{image\textsubscript{3}} through spatial layout analysis and integrates \textit{text\textsubscript{1}} to fortify temporal and statistical reasoning. This holistic approach enables an accurate interpretation of the carbon-free energy distribution depicted in the visual data, which would otherwise remain fragmented under conventional methods. Crucially, this comparison highlights the ``semantic gap'' in standard MLLMs. As shown in the baseline outputs, the model treats visuals as isolated tokens, failing to recognize that \textit{image\textsubscript{2}} serves as the decoding key for the geospatial data in \textit{image\textsubscript{3}}. It merely offers generic descriptions like ``circular charts'' without grasping the underlying logic. In contrast, Pharos-ESG utilizes the injected hierarchical path (e.g., \textit{H3: Net-zero carbon}) as a semantic anchor to link these elements. This transforms the task from simple pattern recognition into logic-grounded reasoning, demonstrating that explicit structure is a prerequisite for interpreting complex financial reports.

\section{Aurora-ESG}


To support downstream multimodal document parsing and financial integration, we build Aurora-ESG via the Pharos-ESG pipeline. To our knowledge, it is the largest structured ESG dataset to date. It collects 3,369 reports and 1,135K content blocks from 2,257 China-listed companies (2021-2025, 41.86\% disclosure rate); 13,057 reports and 4,413K blocks from 2,631 HK-listed companies (2021-2025, full disclosure); and 7,539 reports and 2,261K blocks from 3,769 US-listed companies (2023-2025, 69.99\% disclosure rate).

\paragraph{Hierarchical Composition and Content}
Each ESG report in Aurora-ESG is structured into three levels: \textit{document}, \textit{page}, and \textit{content block}, capturing metadata, page references, and fine-grained semantics, respectively (Table~\ref{tab:Aurora-ESG2}).

\paragraph{Potential Applications} 
(1) \textit{Multimodal Long-Document Reasoning.}  
Aurora-ESG provides ESG documents with full reading order, hierarchical structure, and interleaved multimodal content, supporting the training and evaluation of long-context models on tasks reflecting complex real-world scenarios.
(2) \textit{Cross-Document Consistency and Greenwashing Detection.}  
By aligning content blocks to standardized GRI indicators, Aurora-ESG supports fine-grained cross-document comparisons across companies and years, facilitating detection of disclosure inconsistencies, omissions, and potential greenwashing.
(3) \textit{Investor Sentiment and Decision Impact.}
Aurora-ESG incorporates ESG-specific sentiment annotations, enabling studies on how disclosure tone influences investor perception, trust, and sustainability-related decision-making in behavioral finance contexts.
(4) \textit{Cross-Market ESG Benchmarking.} 
Covering ESG reports from China, Hong Kong, and the U.S., Aurora-ESG offers a unified benchmark for comparing ESG reporting practices across regulatory regimes, supporting structured, cross-jurisdictional analysis of reporting completeness, tone, and structure.

\section{Conclusion}
This study introduced Pharos-ESG, a system for large-scale structured analysis of ESG disclosures. It effectively reconstructs reading order and recovers fragmented semantics, addressing implicit structures from diverse visual layouts. The system also converts visual elements into natural language through a contextualized pipeline. To support financial research, the outputs are enriched with multi-level labels. Experimental evaluations demonstrate that Pharos-ESG outperforms both dedicated document parsing systems and general-purpose multimodal models. In addition, we release Aurora-ESG, a large-scale structured ESG dataset covering disclosures from China, Hong Kong, and the U.S., providing a valuable resource for integrating ESG data into financial analysis and decision-making.

\section{Acknowledgements}
This study was supported by the National Natural Science Foundation of China (NSFC) (72571222 and 72401235), the Natural Science Foundation of Sichuan Province (2024NSFSC1061 and 2025ZNSFSC0041), the Financial Innovation Center of Southwestern University of Finance and Economics (FIC2023E007), and the Artificial Intelligence and Digital Finance Key Laboratory of Sichuan Province.
\bibliographystyle{plainnat}
\bibliography{aaai2026}

\end{document}